\documentclass[letterpaper, 10 pt, conference]{ieeeconf}  
\usepackage{amssymb}
\usepackage{graphicx}
\usepackage{
    amsbsy, amsfonts, amsmath, amssymb, bm, graphicx,
    multicol, mwe, siunitx, subcaption, xcolor, pdfpages, cuted, titlesec, float, booktabs, hyperref, balance
}
\usepackage{stfloats}      
\usepackage{graphicx}
\usepackage{capt-of}

\IEEEoverridecommandlockouts                              

\usepackage{graphics} 
\usepackage{epsfig} 
\usepackage{mathptmx} 
\usepackage{times} 
\usepackage{amsmath} 
\usepackage{amssymb}  
\usepackage{mathtools}

\usepackage{physics}       

\makeatletter
\DeclareRobustCommand\onedot{\futurelet\@let@token\@onedot}
\def\@onedot{\ifx\@let@token.\else.\null\fi\xspace}

\def\ie{\emph{i.e}\onedot}

\makeatother

\title{\LARGE \bf
H\(^2\)-COMPACT: Human–Humanoid Co-Manipulation via Adaptive Contact Trajectory Policies
}

\author{Geeta Chandra Raju Bethala$^{1}$, Hao Huang$^{1}$,  Niraj Pudasaini$^{1}$, Abdullah Mohamed Ali$^{2}$,\\  Shuaihang Yuan$^{1,2}$, Congcong Wen$^{1}$, 
Anthony Tzes$^{2}$, Yi Fang$^{1,2}$%
\thanks{$^{1}$Embodied AI and Robotics Lab, New York University Abu Dhabi, UAE.}%
\thanks{$^{2}$Center for AI and Robotics, New York University Abu Dhabi, UAE.}%
}

\begin{document}
\maketitle
\thispagestyle{empty}
\pagestyle{empty}

\begin{strip}
\vspace{-0.5in}
\begin{center}
\centering
  \includegraphics[width=\textwidth]{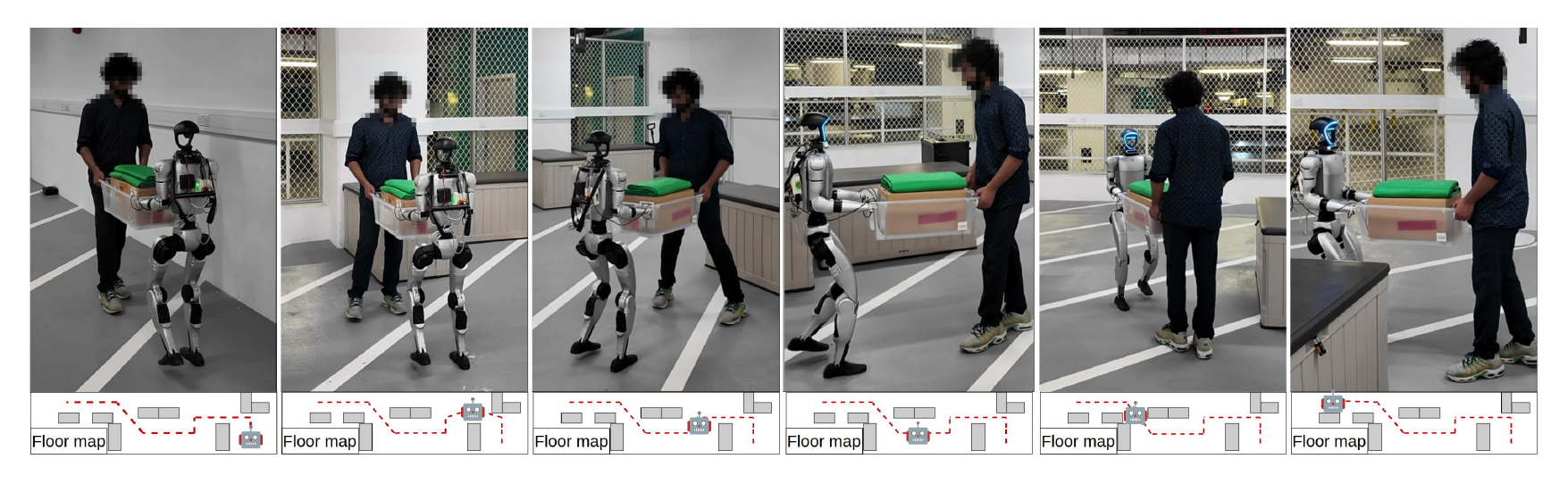}
  \captionof{figure}{Real‐world human–humanoid co‐manipulation. The human leads the humanoid robot—unaware of the route or goal—guiding it via haptic cues to jointly carry the load along the dashed path.}
  \label{robot_comanipulation}
\end{center}
\end{strip}

\begin{abstract}
We present a hierarchical policy‐learning framework that enables a legged humanoid to cooperatively carry extended loads with a human partner using only haptic cues for intent inference. At the upper tier, a lightweight behavior‐cloning network consumes six‐axis force/torque streams from dual wrist‐mounted sensors and outputs whole‐body planar velocity commands that capture the leader’s applied forces. At the lower tier, a deep‐reinforcement‐learning policy—trained under randomized payloads (0–3 kg) and friction conditions in Isaac Gym and validated in MuJoCo and on a real Unitree G1—maps these high‐level twists to stable, under‐load joint trajectories. By decoupling intent interpretation (force → velocity) from legged locomotion (velocity → joints), our method combines intuitive responsiveness to human inputs with robust, load‐adaptive walking. We collect training data without motion‐capture or markers—only synchronized RGB video and F/T readings—employing SAM2 and WHAM to extract 3D human pose and velocity. In real‐world trials, our humanoid achieves cooperative carry‐and‐move performance (completion time, trajectory deviation, velocity synchrony, and follower‐force) on par with a blindfolded human‐follower baseline. This work is the first to demonstrate learned haptic guidance fused with full‐body legged control for fluid human–humanoid co–manipulation. Code and videos are available on \href{https://h2compact.github.io/h2compact/}{https://h2compact.github.io/h2compact/}.
\end{abstract}








\section{INTRODUCTION}
\label{sec:intro}


Current reinforcement learning and imitation learning policies enable humanoid robots to perform highly agile and dynamic actions across diverse terrains \cite{he2024omnih2o}; however, these capabilities typically do not extend to tasks that assist humans in practical daily or workplace activities. Human teams naturally excel at cooperative tasks such as carrying awkward payloads, intuitively sensing and adapting to each other’s applied forces. However, humanoid perception predominantly relies on visual input to interpret and understand the surrounding environment \cite{long2024learninghumanoidlocomotionperceptive,hao2025embodied}, while haptic perception has received limited attention in current research. Thus, achieving similarly fluid collaboration between humans and humanoids remains an open and challenging problem, requiring the humanoid to infer the human’s haptic intent and simultaneously maintain stable locomotion under variable payloads. Previous studies in human–robot co-manipulation have primarily concentrated on fixed-base manipulators \cite{shao2024constraintaware} and wheeled platforms \cite{Mielke_2024}, while most legged-robot controllers either neglect external forces or consider them merely as disturbances rather than as meaningful channels of communication \cite{9684679}. On the contrary, we claim that haptic feedback can instead guide humanoid movements in human-robot co-manipulation.


In this paper, we propose a \textit{hierarchical} policy learning architecture for human-robot co-manipulation that integrates haptic-intent inference with payload-adaptive locomotion on a whole-body humanoid robot. At the higher level, behavioral cloning as our haptic inference model takes force and torque data obtained from a pair of force sensors, generating high-level actions, \ie, whole-body translational and rotational velocity, to interpret human movement intent. At the lower level, a locomotion policy trained using Proximal Policy Optimization (PPO) \cite{schulman2017ppo} translates these high-level actions into low-level actions, \ie, target joint angles of a humanoid. This locomotion policy has been extensively trained in simulated environments with randomized payloads (ranging from 0 to 3 kg) and varying friction conditions within Isaac Gym \cite{makoviychuk2021isaacgymhighperformance}, and subsequently validated in MuJoCo simulations \cite{todorov2012mujoco} and on a real-world humanoid. By decoupling intent interpretation (force to high-level whole-body velocities) from foothold and posture control (high-level whole-body velocities to low-level joint angles), our approach effectively combines responsive haptic feedback with robust locomotion under varying payloads.


To facilitate affordable and widely deployable data collection, we record only an RGB video stream along with two force-torque sensor channels as training data, without using any motion-tracking system. Background clutter is removed using SAM2~\cite{ravi2024sam}, and the resulting images are utilized for 3D human pose and velocity estimation~\cite{shin2024wham}. These estimated human movement data, combined with sensor-measured forces and torques, serve as input for supervised training of our haptic inference model. Our approach is benchmarked against a human–human leader–follower baseline, where the follower is blindfolded. Performance is evaluated using different metrics, including completion time, trajectory deviation, velocity consistency, and average follower-applied force.

\textbf{Our novel contributions} are summarized as follows: 


\begin{enumerate}
    \item  We present a hierarchical policy learning framework to enable humanoids to perform force-adaptive locomotion during physical collaboration with humans.

    \item We introduce a compact and effective haptic inference model trained on minimal sensor data, capable of interpreting human intent through applied forces.

    \item We propose a streamlined, vision-only preprocessing pipeline to accurately estimate human movement without reliance on any additional motion-tracking system or hardware.

    \item We provide a simulation-to-real-world on Unitree G1 humanoid robot~\cite{unitree_g1} using a trained load-adaptive locomotion policy, demonstrating quantitative performance comparable to human–human cooperative tasks.
\end{enumerate}

\section{Related Works}
\label{sec:related}

\noindent \textbf{Physical Human–Robot Co-Manipulation.}
Physical co-manipulation has traditionally targeted fixed-base arms and wheeled platforms, using impedance or admittance schemes to comply with human force inputs \cite{ikeura1995variable}\cite{duchaine2007general}. To disambiguate translation versus rotation in extended‐object carrying, state‐machine heuristics and probabilistic intent estimators \cite{thobbi2011using} have been proposed. Human–human dyad experiments reveal that non-zero interaction forces carry intent and that both lateral and rotational hand motions obey minimum-jerk profiles even when blindfolded \cite{Mielke_2024}. On humanoid arms, proactive haptic transport on Baxter \cite{bussy2012proactive} and EMG- or motion-capture–augmented compliance \cite{peternel2017human}\cite{rozo2016learning} have been demonstrated, but these rely on predefined trajectories or low-DOF manipulators. More recent work explores constraint-aware intent estimation \cite{shao2024constraintaware} and robots taking initiative \cite{rysbek2023robots}, yet none combine learned haptic inference with dynamic whole-body locomotion on a general-purpose humanoid. Crucially, when carrying extended objects a humanoid’s lower body is often occluded —its upper torso remains largely stationary—leaving force feedback as the sole, and most human-natural, modality for inferring the leader’s intent.

\noindent \textbf{RL for Force-Adaptive Humanoid Locomotion.}
Reinforcement learning (RL) has been successfully applied to achieve stable locomotion in quadrupeds \cite{Hwangbo_2019,Lee_2020,tan2018simtoreallearningagilelocomotion} and in humanoids \cite{gu2024humanoid,Singh_2024,wang2025beamdojolearningagilehumanoid}. Compared to classical optimal-control-problem (OCP) based controllers, which must solve high-dimensional trajectory optimizations online—inducing significant latency and exhibiting sensitivity to modeling errors \cite{doi:10.1177/027836498400300206,10286076}—RL-trained policies learn end-to-end control mappings offline and execute in constant time via a single network inference \cite{schulman2017proximalpolicyoptimizationalgorithms,pmlr-v80-haarnoja18b}. The task defined in our paper requires forceful interaction with the human when performing co-manipulation and carrying load while walking. Some of the previous work has learned a policy for tasks requiring forceful interactions with objects, such as dribbling a ball in both quadrupeds \cite{ji2023dribblebotdynamicleggedmanipulation} and humanoids \cite{Haarnoja_2024}, door-opening \cite{zhang2024learningopentraversedoors} and learning control force or compliance at the end effector of quadrupeds with an mounted arm \cite{portela2024learningforcecontrollegged}.

\section{Method}
\label{sec:method}

\begin{figure*}[!t]
  \centering
  \includegraphics[width=\textwidth]{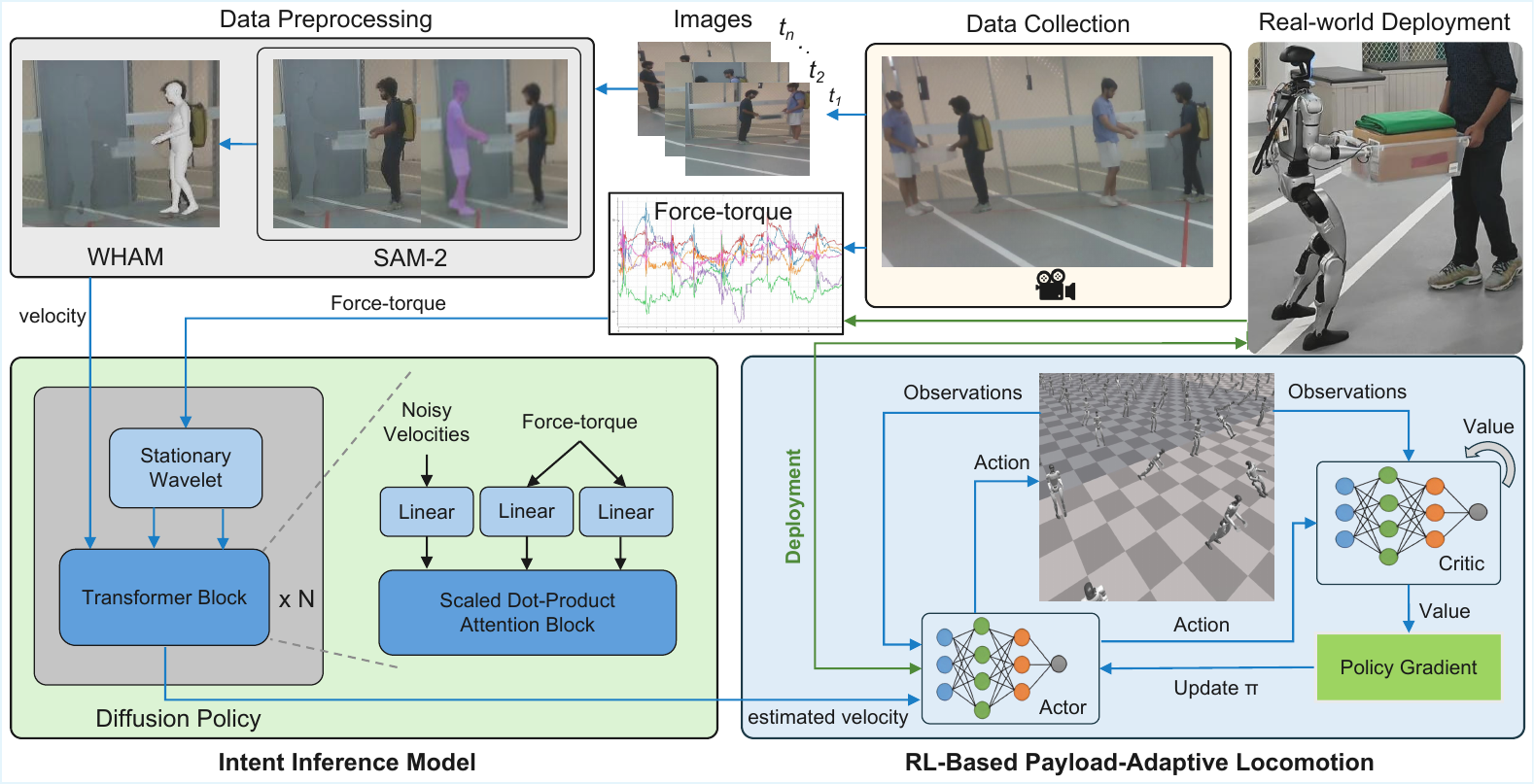}
  \caption{H²-COMPACT's pipeline: raw force/torque and RGB inputs are cleaned by SAM2 and WHAM, then passed through a diffusion-based haptic intent model to generate high-level velocities, which a PPO-trained policy converts into humanoid joint commands.}
  \label{fig:methodology}
\end{figure*}

\subsection{Overview}
\label{subsec:overview}

As shown in Fig.~\ref{fig:methodology}, our co-manipulation framework employs a hierarchical policy learning structure, with each level optimized to address a distinct facet of physical human–robot interaction. At the high level, a behavior cloning policy processes six-axis force and torque data collected from dual ATI sensors mounted on the handles, as depicted in Fig.~\ref{fig:methodology}, directly generating whole-body motion actions—linear velocities $(v_x, v_y)$ and yaw rate $\omega_z$—to capture the human leader's haptic intent. This haptic inference model is trained offline using demonstration data from human–human collaborations, enabling the intent inference model to effectively recognize and respond to subtle force signals inherent in natural cooperative carrying tasks.


These estimated high-level velocities subsequently serve as reference states for a low-level policy trained through deep reinforcement learning. This low-level policy translates whole-body motion actions into accurate target joint angles for precise humanoid robot control. Throughout training, the locomotion policy encounters diverse payloads ranging from 0 to 3 kg, as well as varied friction conditions, enabling it to robustly maintain balance and reliably execute arbitrary whole-body motions across different payload scenarios. By separating the tasks of intent inference (orce to
high-level whole-body velocities) and locomotion control (high-level whole-body velocities to low-level joint
angles), our hierarchical policy architecture achieves both intuitive responsiveness to human inputs and consistent locomotion stability.

\subsection{Intent Inference Model}
\label{subsec:intent}
We design an intent inference model to learn the mapping:
\begin{equation}
   (F,\tau)\;\longmapsto\;\dot{\mathbf{q}}
   = (v_x,v_y,\omega_z),
\end{equation}
where the inputs are the $3$-DoF forces $F\!\in\!\mathbb{R}^{T\times 6}$ and torques $\tau\!\in\!\mathbb{R}^{T\times 6}$ from two ATI Mini-40/45 sensors for the past $T$ force frames up to now, and the output $\dot{\mathbf{q}}\!\in\!\mathbb{R}^{H\times3}$ is the future whole-body planar linear velocity (XY) plus yaw (Z) rate for the next $H$ frames. The pipeline has three stages:
\begin{itemize}
  \item multi-resolution stationary wavelet transform to encode force–torque signals;
  \item a multiscale \emph{conditional diffusion policy} implemented with a block-wise cross-attention Transformer~\cite{chi2023diffusion};
  \item deterministic DDIM sampling~\cite{songdenoising} at inference time.
\end{itemize}
%

\subsubsection{Multi-resolution Force–Torque Encoding}
\label{subsubsec:wavelet}
For every camera frame, \ie, one whole-body velocity data sample, we collect $T$ force–torque samples per sensor, giving a sequence of length $T=HS$. During implementation, we set $H=6$ and $S=33$. After zero-padding the sequence of force-torque to the length of the next power of two, this sequence is decomposed with an $L$-level stationary wavelet transform~\cite{fowler2005redundant}. Note that only the approximation coefficients
\begin{equation}
   \mathcal{A}
   = \bigl\{\,A_\ell\in\mathbb{R}^{T\times6}\bigr\}_{\ell=1}^{L}
\end{equation}
are kept for their robustness to high-frequency noise. Concatenating channels from the two sensors yields
\begin{equation}
   A_{\ell}^{(F)}\in\mathbb{R}^{T\times6},\qquad
   A_{\ell}^{(\tau)}\in\mathbb{R}^{T\times6}.
\end{equation}
Then, $\{A_{\ell}^{(F)}\}_{l=1}^L$ and $\{A_{\ell}^{(\tau)}\}_{l=1}^L$ are reshaped into blocks $(H,S,L,D)$ with $D=6$ before entering the intent inference model. As the ATI sensors collect force-torque data at around $1K$ Hz which is much higher than the recorded video frame rate, by decomposing signals into different frequency bands, stationary wavelets effectively preserve essential low-frequency features while suppressing unwanted high-frequency noise.

\subsubsection{Conditional Diffusion Policy}
\label{subsubsec:diffusion}

\textbf{Forward diffusion in velocity space.}
Given a velocity sequence $y \coloneqq \dot{\mathbf{q}} \!\in\!\mathbb{R}^{H\times3}$, we draw a time step $t\sim\mathcal{U}\{0,\dots,\hat{T}-1\}$ uniformly and get the noisy velocity:
\begin{equation}
   y_t
   = \sqrt{\bar{\alpha}_t}\,y
     + \sqrt{1-\bar{\alpha}_t}\,\varepsilon,
   \qquad
   \varepsilon\sim\mathcal{N}(0,I_4),
\end{equation}
with cosine schedule $\{\beta_t\}_{t=0}^{\hat{T}-1}$ and $\bar{\alpha}_t = \prod_{s=0}^{t}(1-\beta_s)$~\cite{nichol2021improved}.

\textbf{Multiscale Transformer encoder.}
Each horizon query token $q_i\in\mathbb{R}^{d}$ ($d=128$) is obtained by a linear projection of $y_t$ plus a sinusoidal time embedding. For each wavelet level $\ell \in L$, keys and values are learned from the corresponding force–torque approximation coefficients:
\begin{equation}
   k_l = \mathrm{MLP}\!\bigl[\,W_F A^{(F)}_\ell \;||\; W_\tau A^{(\tau)}_\ell\bigr],
   \qquad
   v_\ell = k_\ell,
   \label{eq:kv}
\end{equation}
where $||$ is channel concatenation and $W_f$, $W_\tau$ are learnable weights. Inside an attention module~\cite{vaswani2017attention}, each query attends \emph{block-wise} (all $S$ samples) over multiple wavelet levels, and the final attention matrix is the weighted sum of all attention matrics at each level:
\begin{equation}
   \tilde{A}_{ij}
   = \sum_{\ell=1}^{L} w_{i\ell}\,
     \mathrm{softmax}\!\Bigl(\tfrac{q_i^\top k_{\ell j}}{\sqrt{d}}\Bigr),
   \qquad
   w_{i\ell}\propto
   \exp\!\bigl[-\mathcal{H}(A_{i\ell})\bigr],
\end{equation}
where $\mathcal{H}$ is the Shannon entropy of the level-specific attention matrix. A stack of $n_L=4$ Transformer blocks are defined within the Transformer encoder.

\textbf{Latent stochastic keys.}
To better model the action \textit{distribution}, instead of using a deterministic mapping, each key is sampled from a diagonal Gaussian $\mathcal{N}(\mu,\mathrm{diag}\,\sigma^2)$ where $\mu$ and $\log \sigma^2$ are computed from linear transformations of $k_\ell$ defined in Eq.~\ref{eq:kv}. With the re-parameterisation trick, the KL term in the evidence lower bound is:
\begin{equation}
   \mathcal{L}_{\mathrm{KL}}
   = \tfrac12\,\mathbb{E}\Bigl[
       \norm{\mu}_2^{2}
       + e^{\smash{\log\sigma^{2}}}
       - \log\sigma^{2}
       - 1
     \Bigr].
\end{equation}

\textbf{Training objective.}
The output of the multiscale Transformer encoder $\varepsilon_\theta$ is trained to predict the noise~\cite{ho2020denoising}:
\begin{equation}
   \mathcal{L}_{\mathrm{diff}}
   = \mathbb{E}_{t,y,\varepsilon}
     \bigl[
        \norm{\varepsilon
        -\varepsilon_\theta(y_t,t,\{A^{(F)}_\ell\}_{\ell=1}^L,\{A^{(\tau)}_\ell\}_{\ell=1}^L)}_2^2
     \bigr],
\end{equation}
and the total loss is:
\begin{equation}
   \mathcal{L}
   = \mathcal{L}_{\mathrm{diff}}
     + \lambda_{\mathrm{KL}}\,\mathcal{L}_{\mathrm{KL}},
   \qquad
   \lambda_{\mathrm{KL}} = 10^{-2}.
\end{equation}
We optimize with Adam optimizer by setting learning rate to $10^{-3}$ and batch size to 32.

\subsubsection{Deterministic DDIM Sampling}
\label{subsubsec:ddim}
At test time, we draw an initial Gaussian $y_{\hat{T}}$ and iterate the DDIM update~\cite{songdenoising} for $K=20$ steps:

\begin{align}
   \hat{y}_0
      &= \frac{y_t - \sqrt{1-\bar{\alpha}_t}\,
                  \varepsilon_\theta}{\sqrt{\bar{\alpha}_t}}, \label{eq:ddim_y0} \\
   y_{t-1}
      &= \sqrt{\bar{\alpha}_{t-1}}\;\hat{y}_0
         + \sqrt{1-\bar{\alpha}_{t-1}}\;
           \varepsilon_\theta, \label{eq:ddim_ytm1}
\end{align}
where the output $y_t$ is always conditioned on the multiscale force-torque representation
$\{A^{(F,\tau)}_\ell\}_{\ell=1}^L$. The first horizon token of the final $\hat{y}_0$ is regarded as the high-level actions which are further used as a reference state fed into the low-level policy to generate the target joint angles.

\subsection{RL-Based Payload-Adaptive Locomotion}
\subsubsection{Problem Formulation}
We formulate our payload-adaptive locomotion task as a partially observable Markov decision process (POMDP) and utilize Proximal Policy Optimization (PPO) \cite{schulman2017proximalpolicyoptimizationalgorithms} to learn a parameterized policy \(\pi_\theta(a_t \mid o_{ t})\) that maximizes the expected discounted return while ensuring stable and force-aligned locomotion. We implement PPO from the RSL-RL library \cite{rudin2022learning} with both actor and critic as three‐layer MLPs. 
Mathematically, a POMDP is defined as:
\begin{equation}
\mathcal{M} = \bigl(\,\mathcal{S},\;\mathcal{A},\;T,\;\mathcal{O},\;R,\;\gamma\bigr),
\end{equation}
where a full simulator state \(\mathbf{s}_t\in\mathcal S\) evolves via \(T(s_{t+1}\mid s_t,a_t)\), but at deployment the agent only receives observation \(o_t\in\mathcal O\). Note that the high-level actions generated with Eq.~\ref{eq:ddim_y0} and Eq.~\ref{eq:ddim_ytm1} are observable during both training and deployment. $R$ is the scalar reward function per step, and $\gamma\in[0,1]$ is the discount factor balancing immediate versus long‐term returns. We parameterize a feed-forward policy
\(\pi_\theta(a_t\mid o_t)\) and train it with PPO to maximize
\(\displaystyle J(\theta)=\mathbb{E}_{\pi_\theta}\bigl[\sum_t\gamma^t\,r_t\bigr]\).
To stabilize updates we use the clipped surrogate:
\begin{equation}
\mathcal{L}_{\pi}(\theta)
= \mathbb{E}_t\Bigl[\min\bigl(\rho_t\,\widehat A_t,\,
\mathrm{clip}(\rho_t,1-\epsilon,1+\epsilon)\,\widehat A_t\bigr)\Bigr],
\end{equation}
where
\begin{equation}
\rho_{t}
= \frac{\pi_{\theta}(a_{t}\mid o_{t})}
       {\pi_{\theta_{\text{old}}}(a_{t}\mid o_{t})}
\end{equation}
is the likelihood ratio, \(\epsilon>0\) is the PPO clipping parameter that bounds how far the new policy is allowed to deviate from the old one in each update, and \(\widehat A_{t}\) is an estimate of the advantage function:
\begin{equation}
\widehat A_{t}
= {\sum_{l=0}^{T-t}\gamma^{\,l}r_{t+l}}
- V_{\phi}(o_{t}),
\end{equation}
with \(V_{\phi}(o_{t})\) the value‐function baseline.  

\subsubsection{Training for Load-Adaptation}
%
%
Building on Unitree’s official RL framework, which follows \cite{gu2024humanoid} by using an asymmetric actor‐critic for lower‐body control and phase‐based rewards to learn periodic gaits, we train our policy to handle dynamic payloads by applying randomized external forces at the wrist‐roll link on both hands within Isaac Gym \cite{makoviychuk2021isaacgymhighperformance}. During each episode, each wrist links receive a force in the z-direction sampled from:
\begin{equation}
f \sim \mathcal{U}([-F_{\max},\,F_{\max}]),
\end{equation}
emulating variable payload mass and distribution. We set \(F_{\max}=15\) N to match the robot’s nominal payload capacity, and choose a lower bound of \(-3\) N to allow for small assisting (upward) forces that a human partner might exert when lifting. This randomized force perturbation ensures the learned policy dynamically compensates for both resisting and assisting loads during co‐manipulation.

At each timestep \(t\), the observation vector is
\[
O_t = \bigl[\,
\omega_t,\;
\hat{g_t},\;
C_t,\;
(q_t - q_{0,t}),\;
\dot q_t,\;
a_{t-1},\;
\sin(2\pi \phi_t),\;
\cos(2\pi \phi_t)
\bigr],
\]
where \(\omega_t\) is the body’s angular velocity, \(g_t\) the projected gravity vector, \(C_t\) the randomized $\hat{y}_0$ for linear velocity in x and y directions and angular velocity in the z direction, \(q_t\) and \(\dot q_t\) the 12-DOF joint positions and velocities for legs, \(q_{o,t}\) the default joint positions, and \(\phi_t\) the normalized gait phase. The network then outputs desired legs joint angles \(a_t\) as low-level actions, which are converted to torques by:
\begin{equation}
\tau_{t,i} = K_{p,i}\,(a_{t,i} - a_{0,t,i}) \;-\; K_{d,i}\,\dot a_{t,i},
\end{equation}
where \(i\) indexes each joint. 

\subsubsection{Simulation-to-Real Considerations}
We leverage Isaac Gym for high-speed, GPU-based parallel training of our RL policy and conduct sim-to-sim validation in MuJoCo \cite{todorov2012mujoco}, which is known for its accurate physical dynamics \cite{gu2024humanoid}. This workflow combines the rapid experimentation of Isaac Gym with the high-fidelity but slower CPU-based MuJoCo environment. Finally, the trained policy is deployed and evaluated in sim2real transfer experiments on the physical Unitree G1-23 DOF robot.

To increase robustness to real-world variability, we apply domain randomization throughout training. Specifically, at the start of each episode we randomly perturb the ground contact properties, the robot’s base mass, and apply external pushes.
These randomizations follows the principles of \cite{tobin2017domainrandomizationtransferringdeep} and helps prevent overfitting to any single simulation configuration, resulting a policy that remains stable under variations in surface friction, load, and unexpected collisions in both sim-to-sim and sim-to-real transfer.

\subsection{Data Preprocessing}

All sensor streams—force and torque from the two ATI Mini45 units and RGB video—are time‐stamped in ROS and aligned in post‐processing to ensure synchronization.  This unified timeline allows us to associate each image frame with the corresponding six‐axis wrench readings on both handles.

\subsubsection{Image Cleaning via SAM2}  
\label{subsubsec:sam2}
To make pose estimation robust in cluttered or crowded scenes, we first apply the Segment Anything Model v2 (SAM2) \cite{ravi2024sam} to each RGB frame.  SAM2 produces binary masks for all non‐human objects and extraneous persons.  We then inpaint these masked regions by copying the original background texture, effectively “erasing’’ occluders while preserving the true scene context.  This preprocessing step ensures that downstream pose estimators operate on clean, human‐only imagery, even in busy environments.

\subsubsection{Human Pose and Velocity Estimation with WHAM}  
\label{subsubsec:wham}
Cleaned frames are passed to the WHAM 3D human pose‐and‐velocity network \cite{shin2024wham}, which outputs joint positions in a \textit{world} frame at each timestamp.  We compute linear and angular velocities by finite‐difference of the torso joint over consecutive frames and then convert them to \textit{local} frame centered on humans:
\begin{equation}
  v_h(t) \;=\; \frac{p_h(t) - p_h(t - \Delta t)}{\Delta t}
  \quad,\quad
  \omega_h(t) \;=\; \frac{\theta_h(t) - \theta_h(t - \Delta t)}{\Delta t}\,.
\end{equation}
Here \(p_h\) and \(\theta_h\) denote the 3D position and orientation of the human’s movement pose, and \(\Delta t\) is the frame interval.

\subsubsection{Fusion for Model Training}  
\label{subsubsec:fuse}
Finally, we pair each \( \{v_h(t),\omega_h(t)\}\) sample with the synchronized six‐axis force/torque readings from both handles, yielding a compact, sensor‐agnostic dataset. This fusion of haptic and kinematic streams forms the basis for training our co‐manipulation intent inference model as described above, without any specialized motion capture system or hardware.

\section{Experiment}
\label{sec:experiment}

\subsection{Hardware Design}

\begin{figure*}[t]
  \centering
    \centering
    \includegraphics[width=\textwidth]{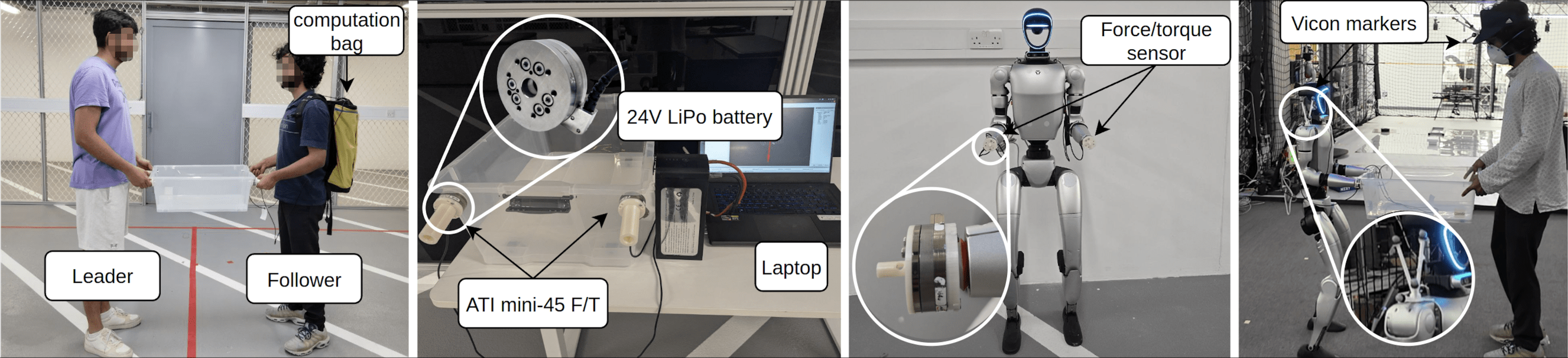}
    \caption{Overview of experimental hardware. From left to right: \textbf{(a)} leader–follower rig with follower’s computation backpack; \textbf{(b)} custom instrumented box, 3D-printed handles and ATI Mini45 F/T sensors; \textbf{(c)} on-robot deployment, ATI Mini45 sensors mounted on the Unitree G1 wrists; \textbf{(d)} custom Vicon marker mounts for motion-capture evaluation.}
    \label{data_robot}
\end{figure*}

We modified a box for data-collection by adding 3D-printed handles to mount an ATI Mini45 six-axis force/torque sensor via a bespoke adapter Fig.~\ref{data_robot}.b. The sensor was interfaced through an EtherCAT OEM F/T module \cite{ati2024mini45} and powered by a 24 V LiPo battery. Force/torque signals were sampled at 1 kHz and streamed over USB into a laptop running ROS, where they were time synchronized with image data and recorded into rosbag \cite{inproceedings}.

For on-robot deployment Fig.~\ref{data_robot}.c, the ATI Mini45 was affixed to the Unitree G1 wrist using custom 3D-printed mounts. It drew power from the robot’s internal 24V bus, and all sensor data were transmitted via EtherCAT to the onboard computer \cite{unitree2025g1developer}.

\subsection{Data Collection}

\begin{figure*}[t]
  \centering
    \centering
    \includegraphics[width=\textwidth]{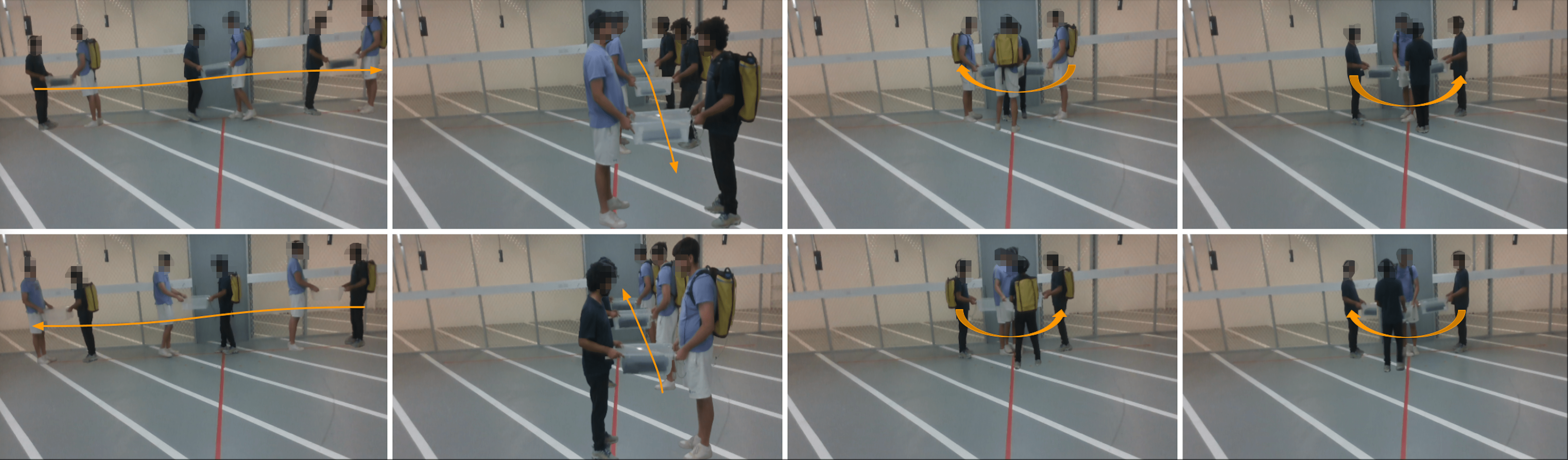}
    \caption{The eight motion primitives executed during dyadic data collection}
    \label{data_collection}
\end{figure*}

We conducted dyadic leader–follower trials Fig.~\ref{data_collection}. In each run, the leader and blindfolded follower grasp opposite handles of the instrumented box. The follower’s backpack streams F/T data at 1 kHz into ROS, while a fixed RGB camera captures the full scene. All streams are timestamped and logged together.

Eight motion primitives were defined: pure translations (forward, backward, left, right), leader-centric rotations (clockwise, counterclockwise), and follower-centric rotations (clockwise, counterclockwise). Each primitive was repeated three times under four payloads (0, 1, 3, 4 kg), for a total of 96 trials. Blindfolding the follower and withholding trajectory cues ensured reliance solely on haptic communication.

\subsection{Real world testing with Humanoid robot}

We evaluated human–humanoid co‐manipulation in two settings: first in an open‐area course with obstacles requiring on‐the‐fly avoidance Fig.~\ref{robot_comanipulation}, and then in our Vicon motion‐capture arena Fig.~\ref{data_robot}.d. In each trial a human leader guided the Unitree G1—unaware of the route—while jointly carrying a box whose payload varied between 0 and 5 kg. No visual or verbal cues were provided to the robot: all guidance came through the wrist‐mounted ATI sensors. Three participants performed ten repetitions of the full carry‐and‐move task in each environment.  

To quantify co–manipulation performance we adopt four metrics inspired from prior dyadic studies \cite{Mielke_2024,iros_comanip_flex}.

\noindent \textbf{Completion Time.}  
We denote by $T_c$ the elapsed time from the onset of cooperative motion ($t_s$) to task completion ($t_e$). We detect $t_s$ when the object’s centre of mass first moves beyond 5\% of the total displacement, and $t_e$ when it remains within 95\% of the goal for at least 0.5s. Formally,
\[
  T_c = t_e - t_s
  \label{eq:completion_time}
\]
\emph{What it is:} The total carry-and-move duration.  
\emph{Why it matters:} A longer $T_c$ may indicate misunderstanding of haptic cues and poor cooperation with the leader.

\noindent \textbf{Mean Trajectory Deviation.}  
We quantify spatial agreement by averaging the Euclidean distance between human and robot end-effector positions:
\begin{equation}
  \Delta_{\rm traj}
  = \frac{1}{t_e - t_s}
    \int_{t_s}^{t_e}
    \bigl\|x_h(t) - x_r(t)\bigr\| \,\mathrm{d}t
  \label{eq:traj_dev}
\end{equation}
where $x_h(t),x_r(t)\in\mathbb{R}^3$ are the human and robot trajectories.  
\emph{What it is:} The average path mismatch.  
\emph{Why it matters:} A low $\Delta_{\rm traj}$ indicates the robot is moving intuitively alongside the human; large values reveal mis-coordination.

\noindent \textbf{Mean Velocity Difference.}  
We measure kinematic synchrony by the average speed mismatch:
\begin{equation}
  \Delta_{v}
  = \frac{1}{t_e - t_s}
    \int_{t_s}^{t_e}
    \bigl\|\dot x_h(t) - \dot x_r(t)\bigr\| \,\mathrm{d}t
  \label{eq:vel_diff}
\end{equation}
with $\dot x_h,\dot x_r\in\mathbb{R}^3$ denoting instantaneous velocities.  
\emph{What it is:} The mean absolute difference in speed.  
\emph{Why it matters:} Even if positions align, jerky or lagging velocity profiles are uncomfortable and signal poor coupling.

\noindent \textbf{Average Follower Force.}  
We define the mean follower force $\bar F$ as:
\begin{equation}
  \bar F
  = \frac{1}{t_e - t_s}
    \int_{t_s}^{t_e}
    \bigl(\|f_1(t)\| + \|f_2(t)\|\bigr)\,\mathrm{d}t
  \label{eq:force}
\end{equation}
where $f_1(t),f_2(t)\in\mathbb{R}^3$ are the instantaneous force vectors measured by the two ATI sensors.  
\emph{What it is:} The average total magnitude of force the follower (human or robot) senses from the leader’s haptic inputs.  
\emph{Why it matters:} A lower $\bar F$ indicates the follower is highly responsive—interpreting the leader’s intent and moving with minimal force—while high values imply the leader must push harder to drive cooperation.

All kinematic data were recorded with a Vicon motion capture system and a custom 3D-printed marker mount \cite{10977589}.  We benchmark human–humanoid performance against a human–human baseline collected using three dyads (one leader, one blindfolded follower) each performing ten trials of box transport under varying load conditions.  

\section{Result}
\label{sec:result}

\begin{figure}[]
  \includegraphics[width=\columnwidth]{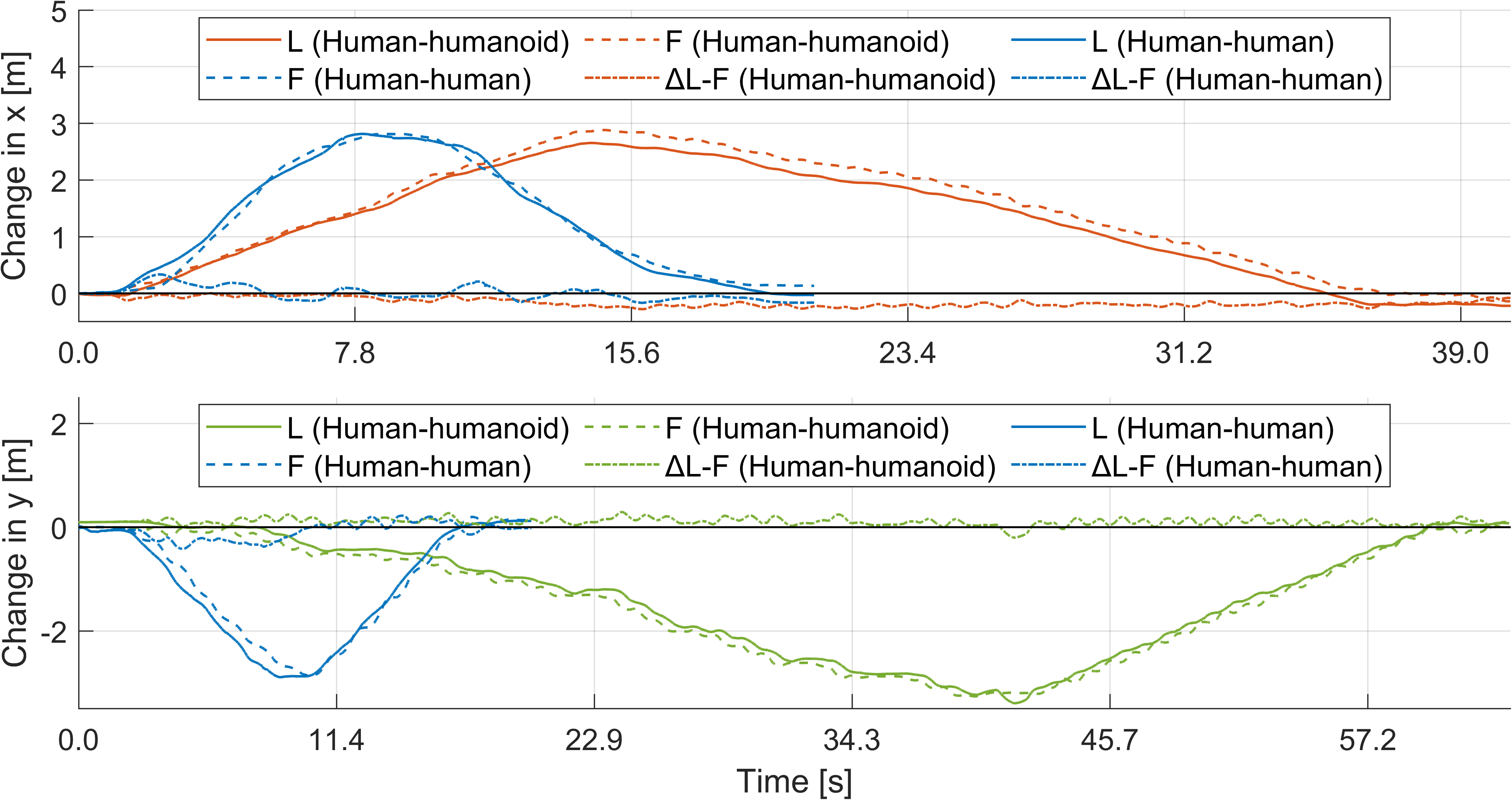}
 \caption{Position tracking of the humanoid and human followers relative to the leader in both scenarios, along with the corresponding position error between the leader and each follower. \textbf{Top:} Motion along the x-axis. \textbf{Bottom:} Motion along the y-axis.}
  \label{fig:position}
\end{figure}

\begin{figure}[]
  \includegraphics[width=\columnwidth]{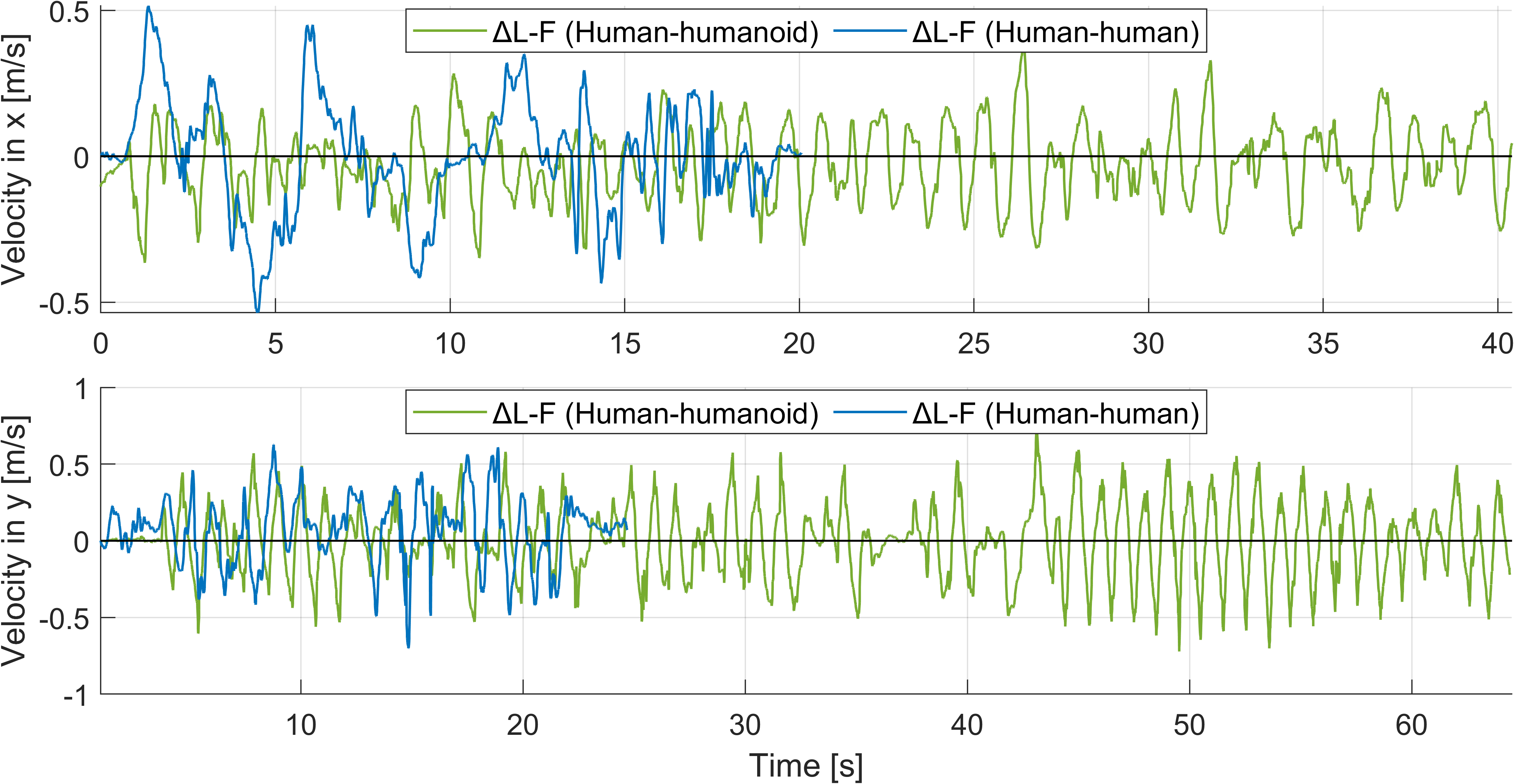}
    \caption{Velocity error tracking between the leader and the corresponding humanoid or human follower in each case. \textbf{Top:} Tracking along the x-axis. \textbf{Bottom:} Tracking along the y-axis.}
  \label{fig:velocity}
\end{figure}

\begin{table}[t]
  \centering
  \caption{Comparison of evaluation metrics between human–human and human–humanoid co–manipulation (mean).}
  \label{tab:metrics_comparison}
  \setlength{\tabcolsep}{2pt}
  \begin{tabular}{lcc}
    \toprule
    \textbf{Metric} & \textbf{Human–Human} & \textbf{Human–Humanoid} \\
    \midrule
    Completion Time $T_c$ (s) $\downarrow$  
      & 23.78  
      & 51.47 \\

    Trajectory Deviation $\Delta_{\rm traj}$ (m) $\downarrow$  
      & 0.1109  
      & 0.1294 \\

    Velocity Difference $\Delta_v$ (m/s) $\downarrow$  
      & 0.165  
      & 0.143 \\

    Average Follower Force $\bar F$ (N) $\downarrow$ 
      & 17.355 
      & 16.230 \\
    \bottomrule
  \end{tabular}
\end{table}

\subsection{Human-humanoid co-manipulation}

Table \ref{tab:metrics_comparison} summarizes the four metrics for human–human versus human–humanoid dyads. Fig.~\ref{fig:position} plots representative $x$– and $y$–position traces, while Fig.~\ref{fig:velocity} shows the corresponding velocity profiles.

\noindent \textbf{Completion Time.}
Human–humanoid pairs took longer ($T_c=51.47,$s) than human–human ($23.78,$s). This is attributable not to poor cooperation but to the G1’s 0.8 m/s speed cap—in spite of accurate haptic following, the robot simply cannot match a human sprint, as confirmed by the smooth, human‐like trajectory in Fig.~\ref{robot_comanipulation}.

\noindent \textbf{Trajectory Deviation.}
Mean deviation $\Delta_{\rm traj}$ rose only slightly from 0.1109 m (human–human) to 0.1294 m (human–humanoid), indicating that the robot’s path remained tightly coupled to the leader Fig.~\ref{fig:position}.

\noindent \textbf{Velocity Difference.}
The average speed mismatch $\Delta_v$ actually decreased (0.165 m/s → 0.143 m/s), showing improved kinematic synchrony in the humanoid case Fig.~\ref{fig:velocity}.

\noindent \textbf{Average Follower Force.}
Mean force $\bar F$ fell from 17.36 N to 16.23 N when the robot followed, demonstrating reduced haptic effort by the human Table \ref{tab:metrics_comparison}.

Together these results confirm that, despite hardware speed limits, the load‐adaptive G1 policy yields cooperative, stable co‐manipulation on par with—and in some respects exceeding—pure human dyads marginally.

\subsection{Load-Adaptive RL policy}

\begin{figure}[h!]
  \centering
  \includegraphics[width=\columnwidth]{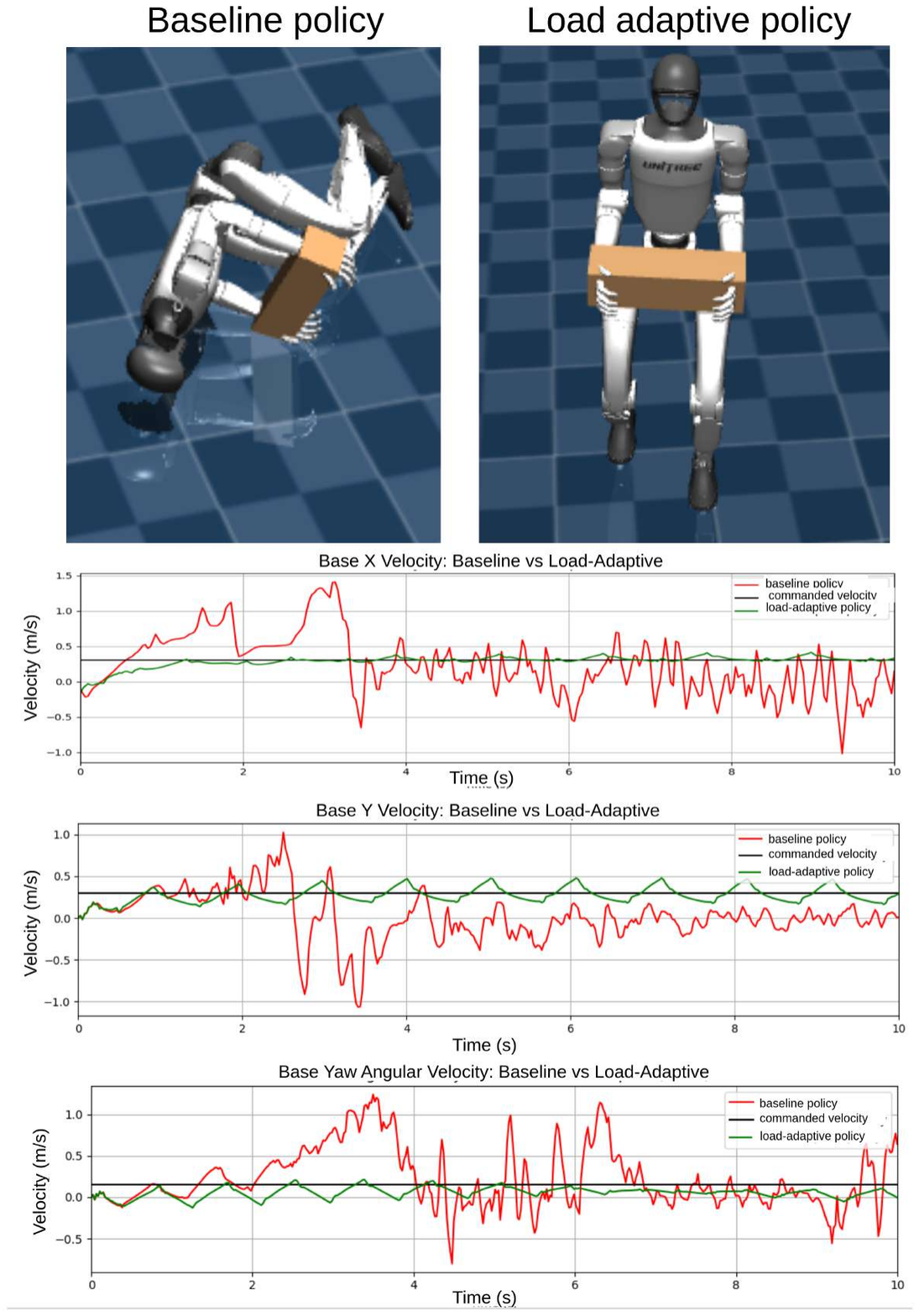}
  \caption{Sim2Sim transfer of the baseline and load‐adaptive policies in MuJoCo under a constant 30\,N payload. \textbf{Top:} The baseline policy tips and falls (left) whereas the load‐adaptive policy holds the box stably (right). \textbf{Bottom:} Commanded vs.\ actual base x‐velocity, y‐velocity, and yaw angular velocity rate over 0–10\,s for both controllers (red dotted: baseline; green solid: load‐adaptive), showcasing the adaptive policy’s smooth, accurate tracking under load.}
  \label{fig:sim2sim_transfer}
\end{figure}

We tested the policy as a sim2sim transfer to the MuJoCo simulator and compared the base linear velocities (along both x- and y-axes) and yaw rate to verify the training under our use cases.
In every trial the robot carried a constant 30 N payload. The plots shows the commanded versus actual x-velocity (first) and y-velocity (second), and the bottom row shows the yaw rate for both the baseline policy (labeled red) and our load-adaptive policy (labeled green).

As shown in the bottom panel of Fig.~\ref{fig:sim2sim_transfer} under the applied load, the baseline controller exhibits large overshoots, oscillations, and even reversals in both linear channels—behavior that ultimately leads to instability (as seen in the fallen pose on the left of first row) top panel of Fig.~\ref{fig:sim2sim_transfer} . By contrast, the load-adaptive policy tracks the constant command almost perfectly across all three channels, with minimal transient error and smooth convergence to the target velocity as seen in bottom panel of Fig.~\ref{fig:sim2sim_transfer}, demonstrating substantially improved stability and payload robustness in sim2sim transfer.

\section{CONCLUSION}
\label{sec:conlc}
In this work, we introduced a two‐stage learning framework that marries haptic‐intent inference with payload‐adaptive locomotion to enable fluid human–humanoid co‐manipulation. A diffusion‐based behavior‐cloning model translates dual‐wrist force/torque readings into whole‐body velocity commands, while a PPO‐trained policy on a simulated Unitree G1 learns to convert these high‐level twists into stable, under‐load joint trajectories. By decoupling intent interpretation from foothold and posture control, our approach achieves close leader–follower coupling—in some metrics even outperforming blindfolded human dyads—despite the robot’s 0.8 m/s speed cap. We further demonstrated reliable sim2sim transfer in MuJoCo and zero‐shot sim2real deployment in both open‐area and Vicon‐tracked obstacle courses, validating robust, intuitive cooperation under varying payloads. Our results highlight the promise of combining haptic feedback with deep reinforcement learning for everyday humanoid assistance tasks.

{
    \small
    \bibliographystyle{ieeetr}

    \bibliography{main}

\begin{thebibliography}{10}

\bibitem{he2024omnih2o}
T.~He, Z.~Luo, X.~He, W.~Xiao, C.~Zhang, W.~Zhang, K.~Kitani, C.~Liu, and G.~Shi, ``Omnih2o: Universal and dexterous human-to-humanoid whole-body teleoperation and learning,'' {\em arXiv preprint arXiv:2406.08858}, 2024.

\bibitem{long2024learninghumanoidlocomotionperceptive}
J.~Long, J.~Ren, M.~Shi, Z.~Wang, T.~Huang, P.~Luo, and J.~Pang, ``Learning humanoid locomotion with perceptive internal model,'' 2024.

\bibitem{hao2025embodied}
Y.~Hao, G.~C.~R. Bethala, N.~Pudasaini, H.~Huang, S.~Yuan, C.~Wen, B.~Huang, A.~Nguyen, and Y.~Fang, ``Embodied chain of action reasoning with multi-modal foundation model for humanoid loco-manipulation,'' {\em arXiv preprint arXiv:2504.09532}, 2025.

\bibitem{shao2024constraintaware}
Y.~Shao, T.~Li, S.~Keyvanian, P.~Chadhuari, V.~Kumar, and N.~Figueroa, ``Constraint-aware intent estimation for dynamic human-robot object co-manipulation,'' in {\em Robotics: Science and Systems}, 2024.

\bibitem{Mielke_2024}
E.~Mielke, E.~Townsend, D.~Wingate, J.~L. Salmon, and M.~D. Killpack, ``Human-robot planar co-manipulation of extended objects: data-driven models and control from human-human dyads,'' {\em Frontiers in Neurorobotics}, vol.~18, 2024.

\bibitem{9684679}
Y.~Ma, F.~Farshidian, T.~Miki, J.~Lee, and M.~Hutter, ``Combining learning-based locomotion policy with model-based manipulation for legged mobile manipulators,'' {\em IEEE Robotics and Automation Letters}, vol.~7, no.~2, pp.~2377--2384, 2022.

\bibitem{schulman2017ppo}
J.~Schulman, F.~Wolski, P.~Dhariwal, A.~Radford, and O.~Klimov, ``Proximal policy optimization algorithms,'' {\em arXiv preprint arXiv:1707.06347}, 2017.

\bibitem{makoviychuk2021isaacgymhighperformance}
V.~Makoviychuk, L.~Wawrzyniak, Y.~Guo, M.~Lu, K.~Storey, M.~Macklin, D.~Hoeller, N.~Rudin, A.~Allshire, A.~Handa, and G.~State, ``Isaac gym: High performance gpu-based physics simulation for robot learning,'' 2021.

\bibitem{todorov2012mujoco}
E.~Todorov, T.~Erez, and Y.~Tassa, ``Mujoco: A physics engine for model-based control,'' in {\em 2012 IEEE/RSJ International Conference on Intelligent Robots and Systems}, pp.~5026--5033, IEEE, 2012.

\bibitem{ravi2024sam}
N.~Ravi, V.~Gabeur, Y.-T. Hu, R.~Hu, C.~Ryali, T.~Ma, H.~Khedr, R.~R{\"a}dle, C.~Rolland, L.~Gustafson, {\em et~al.}, ``Sam 2: Segment anything in images and videos,'' {\em arXiv preprint arXiv:2408.00714}, 2024.

\bibitem{shin2024wham}
S.~Shin, J.~Kim, E.~Halilaj, and M.~J. Black, ``Wham: Reconstructing world-grounded humans with accurate 3d motion,'' in {\em Proceedings of the IEEE/CVF Conference on Computer Vision and Pattern Recognition}, pp.~2070--2080, 2024.

\bibitem{unitree_g1}
U.~Robotics, ``Unitree g1 humanoid robot.'' \url{https://www.unitree.com/g1}, 2024.
\newblock Accessed: 2025-05-16.

\bibitem{ikeura1995variable}
R.~Ikeura and H.~Inooka, ``Variable impedance control of a robot for cooperation with a human,'' in {\em Proceedings of 1995 IEEE International Conference on Robotics and Automation}, vol.~3, pp.~3097--3102 vol.3, 1995.

\bibitem{duchaine2007general}
V.~Duchaine and C.~M. Gosselin, ``General model of human-robot cooperation using a novel velocity based variable impedance control,'' in {\em Second Joint EuroHaptics Conference and Symposium on Haptic Interfaces for Virtual Environment and Teleoperator Systems (WHC'07)}, pp.~446--451, 2007.

\bibitem{thobbi2011using}
A.~Thobbi, Y.~Gu, and W.~Sheng, ``Using human motion estimation for human-robot cooperative manipulation,'' in {\em 2011 IEEE/RSJ International Conference on Intelligent Robots and Systems}, pp.~2873--2878, 2011.

\bibitem{bussy2012proactive}
A.~Bussy, P.~Gergondet, A.~Kheddar, F.~Keith, and A.~Crosnier, ``Proactive behavior of a humanoid robot in a haptic transportation task with a human partner,'' in {\em 2012 IEEE RO-MAN: The 21st IEEE International Symposium on Robot and Human Interactive Communication}, pp.~962--967, 2012.

\bibitem{peternel2017human}
L.~Peternel, N.~Tsagarakis, and A.~Ajoudani, ``A human–robot co-manipulation approach based on human sensorimotor information,'' {\em IEEE Transactions on Neural Systems and Rehabilitation Engineering}, vol.~25, no.~7, pp.~811--822, 2017.

\bibitem{rozo2016learning}
L.~Rozo, S.~Calinon, D.~Caldwell, P.~Jimenez, and C.~Torras, ``Learning physical collaborative robot behaviors from human demonstrations,'' {\em IEEE Transactions on Robotics}, vol.~32, pp.~1--15, 04 2016.

\bibitem{rysbek2023robots}
Z.~Rysbek, K.-H. Oh, A.~Shervedani, T.~Klemencic, M.~Zefran, and B.~Di~Eugenio, ``Robots taking initiative in collaborative object manipulation: Lessons from physical human-human interaction,'' 04 2023.

\bibitem{Hwangbo_2019}
J.~Hwangbo, J.~Lee, A.~Dosovitskiy, D.~Bellicoso, V.~Tsounis, V.~Koltun, and M.~Hutter, ``Learning agile and dynamic motor skills for legged robots,'' {\em Science Robotics}, vol.~4, Jan. 2019.

\bibitem{Lee_2020}
J.~Lee, J.~Hwangbo, L.~Wellhausen, V.~Koltun, and M.~Hutter, ``Learning quadrupedal locomotion over challenging terrain,'' {\em Science Robotics}, vol.~5, Oct. 2020.

\bibitem{tan2018simtoreallearningagilelocomotion}
J.~Tan, T.~Zhang, E.~Coumans, A.~Iscen, Y.~Bai, D.~Hafner, S.~Bohez, and V.~Vanhoucke, ``Sim-to-real: Learning agile locomotion for quadruped robots,'' 2018.

\bibitem{gu2024humanoid}
X.~Gu, Y.-J. Wang, and J.~Chen, ``Humanoid-gym: Reinforcement learning for humanoid robot with zero-shot sim2real transfer,'' {\em arXiv preprint arXiv:2404.05695}, 2024.

\bibitem{Singh_2024}
R.~P. Singh, M.~Morisawa, M.~Benallegue, Z.~Xie, and F.~Kanehiro, ``Robust humanoid walking on compliant and uneven terrain with deep reinforcement learning,'' in {\em 2024 IEEE-RAS 23rd International Conference on Humanoid Robots (Humanoids)}, p.~497–504, IEEE, Nov. 2024.

\bibitem{wang2025beamdojolearningagilehumanoid}
H.~Wang, Z.~Wang, J.~Ren, Q.~Ben, T.~Huang, W.~Zhang, and J.~Pang, ``Beamdojo: Learning agile humanoid locomotion on sparse footholds,'' 2025.

\bibitem{doi:10.1177/027836498400300206}
H.~Miura and I.~Shimoyama, ``Dynamic walk of a biped,'' {\em The International Journal of Robotics Research}, vol.~3, no.~2, pp.~60--74, 1984.

\bibitem{10286076}
P.~M. Wensing, M.~Posa, Y.~Hu, A.~Escande, N.~Mansard, and A.~D. Prete, ``Optimization-based control for dynamic legged robots,'' {\em IEEE Transactions on Robotics}, vol.~40, pp.~43--63, 2024.

\bibitem{schulman2017proximalpolicyoptimizationalgorithms}
J.~Schulman, F.~Wolski, P.~Dhariwal, A.~Radford, and O.~Klimov, ``Proximal policy optimization algorithms,'' 2017.

\bibitem{pmlr-v80-haarnoja18b}
T.~Haarnoja, A.~Zhou, P.~Abbeel, and S.~Levine, ``Soft actor-critic: Off-policy maximum entropy deep reinforcement learning with a stochastic actor,'' in {\em Proceedings of the 35th International Conference on Machine Learning} (J.~Dy and A.~Krause, eds.), vol.~80 of {\em Proceedings of Machine Learning Research}, pp.~1861--1870, PMLR, 10--15 Jul 2018.

\bibitem{ji2023dribblebotdynamicleggedmanipulation}
Y.~Ji, G.~B. Margolis, and P.~Agrawal, ``Dribblebot: Dynamic legged manipulation in the wild,'' 2023.

\bibitem{Haarnoja_2024}
T.~Haarnoja, B.~Moran, G.~Lever, S.~H. Huang, D.~Tirumala, J.~Humplik, M.~Wulfmeier, S.~Tunyasuvunakool, N.~Y. Siegel, R.~Hafner, M.~Bloesch, K.~Hartikainen, A.~Byravan, L.~Hasenclever, Y.~Tassa, F.~Sadeghi, N.~Batchelor, F.~Casarini, S.~Saliceti, C.~Game, N.~Sreendra, K.~Patel, M.~Gwira, A.~Huber, N.~Hurley, F.~Nori, R.~Hadsell, and N.~Heess, ``Learning agile soccer skills for a bipedal robot with deep reinforcement learning,'' {\em Science Robotics}, vol.~9, Apr. 2024.

\bibitem{zhang2024learningopentraversedoors}
M.~Zhang, Y.~Ma, T.~Miki, and M.~Hutter, ``Learning to open and traverse doors with a legged manipulator,'' 2024.

\bibitem{portela2024learningforcecontrollegged}
T.~Portela, G.~B. Margolis, Y.~Ji, and P.~Agrawal, ``Learning force control for legged manipulation,'' 2024.

\bibitem{chi2023diffusion}
C.~Chi, Z.~Xu, S.~Feng, E.~Cousineau, Y.~Du, B.~Burchfiel, R.~Tedrake, and S.~Song, ``Diffusion policy: Visuomotor policy learning via action diffusion,'' {\em International Journal of Robotics Research}, p.~02783649241273668, 2023.

\bibitem{songdenoising}
J.~Song, C.~Meng, and S.~Ermon, ``Denoising diffusion implicit models,'' in {\em International Conference on Learning Representations}, 2021.

\bibitem{fowler2005redundant}
J.~E. Fowler, ``The redundant discrete wavelet transform and additive noise,'' {\em IEEE Signal Processing Letters}, vol.~12, no.~9, pp.~629--632, 2005.

\bibitem{nichol2021improved}
A.~Q. Nichol and P.~Dhariwal, ``Improved denoising diffusion probabilistic models,'' in {\em International conference on machine learning}, pp.~8162--8171, PMLR, 2021.

\bibitem{vaswani2017attention}
A.~Vaswani, N.~Shazeer, N.~Parmar, J.~Uszkoreit, L.~Jones, A.~N. Gomez, {\L}.~Kaiser, and I.~Polosukhin, ``Attention is all you need,'' {\em Advances in Neural Information Processing Systems}, vol.~30, 2017.

\bibitem{ho2020denoising}
J.~Ho, A.~Jain, and P.~Abbeel, ``Denoising diffusion probabilistic models,'' {\em Advances in Neural Information Processing Systems}, vol.~33, pp.~6840--6851, 2020.

\bibitem{rudin2022learning}
N.~Rudin, D.~Hoeller, P.~Reist, and M.~Hutter, ``Learning to walk in minutes using massively parallel deep reinforcement learning,'' in {\em Proceedings of the 5th Conference on Robot Learning}, vol.~164 of {\em Proceedings of Machine Learning Research}, pp.~91--100, PMLR, 2022.

\bibitem{tobin2017domainrandomizationtransferringdeep}
J.~Tobin, R.~Fong, A.~Ray, J.~Schneider, W.~Zaremba, and P.~Abbeel, ``Domain randomization for transferring deep neural networks from simulation to the real world,'' 2017.

\bibitem{ati2024mini45}
{ATI Industrial Automation, Inc.}, ``Mini45 titanium six‐axis force/torque sensor,'' 2024.
\newblock Accessed: 2025-04-28.

\bibitem{inproceedings}
M.~Quigley, K.~Conley, B.~Gerkey, J.~Faust, T.~Foote, J.~Leibs, R.~Wheeler, and A.~Ng, ``Ros: an open-source robot operating system,'' 01 2009.

\bibitem{unitree2025g1developer}
{Unitree Robotics, Inc.}, ``G1 developer documentation,'' 2025.
\newblock Accessed: 2025-04-28.

\bibitem{iros_comanip_flex}
D.~Sirintuna, A.~Giammarino, and A.~Ajoudani, ``Human-robot collaborative carrying of objects with unknown deformation characteristics,'' in {\em 2022 IEEE/RSJ International Conference on Intelligent Robots and Systems (IROS)}, pp.~10681--10687, 2022.

\bibitem{10977589}
M.~Hamandi, A.~M. Ali, N.~Evangeliou, A.~Tzes, and F.~Khorrami, ``Generating distinctive marker configurations for robot detection in motion capture systems,'' in {\em 2025 11th International Conference on Automation, Robotics, and Applications (ICARA)}, pp.~247--251, 2025.

\end{thebibliography}
    
}

\newpage
\newpage
\section*{APPENDIX}
\label{sec:appendix}


\begin{table}[H]
  \centering
  \caption{Hyperparameter Settings}
  \label{tab:hyperparameters}
  \begin{tabular}{l c}
    \toprule
    \multicolumn{2}{c}{\textbf{Environment \& Rollout}} \\
    \midrule
    Number of Environments                  & 4096                  \\
    Episode Length (rollout steps per env)  & 4000         \\
    \midrule
    \multicolumn{2}{c}{\textbf{PPO Training}} \\
    \midrule
    Clip Parameter ($\epsilon$)             & 0.2                   \\
    Discount Factor ($\gamma$)              & 0.998                 \\
    GAE Lambda ($\lambda$)                  & 0.95                  \\
    Value-loss Coefficient ($c_{1}$)        & 1.0                   \\
    Entropy Coefficient ($c_{2}$)           & 0.0                   \\
    Learning Rate                           & $1\times10^{-3}$      \\
    Max Gradient Norm                       & 1.0                   \\
    Value Loss Clipping                     & enabled               \\
    KL-divergence Target ($\delta_{\mathrm{KL}}$) & 0.01          \\
    LR Schedule                             & fixed                 \\
    \midrule
    \multicolumn{2}{c}{\textbf{Observation}} \\
    \midrule
    Actor Obs.\ Dim.                        & 47                    \\
    Critic (Privileged) Obs.\ Dim.          & 50                    \\
    \bottomrule
  \end{tabular}
\end{table}

\begin{table}[H]
  \centering
  \caption{Domain Randomization}
  \label{tab:domain_rand}
  \begin{tabular}{l c}
    \toprule
    \textbf{Parameter}                  & \textbf{Value}       \\
    \midrule
    Randomize friction                  & True                 \\
    Friction range                      & [0.1,\,1.25]         \\
    Randomize base mass                 & True                 \\
    Added mass range                    & [$-$1.0,\,3.0]         \\
    Push robots                         & True                 \\
    Push interval (s)                   & 5                    \\
    Max push velocity (m/s)             & 1.5                  \\
    \bottomrule
  \end{tabular}
\end{table}

\begin{table}[H]
  \centering
  \caption{Reward components, expression and scale}
  \label{tab:rewards_refined}
  \begin{tabular}{@{}l l c@{}}
    \toprule
    \textbf{Component} & \textbf{Expression} & \textbf{Scale ($\mu$)} \\
    \midrule
    Z‐axis linear‐velocity penalty
      & $r_{\mathrm{linVelZ}} = (\dot P^b_{z})^{2}$
      & $-2.0$ \\
    XY angular‐velocity penalty
      & $r_{\mathrm{angVelXY}} = \sum_{k=x,y}(\omega^b_{k})^{2}$
      & $-0.05$ \\
    Base‐orientation penalty
      & $r_{\mathrm{orient}} = \sum_{k=x,y}(g^b_{k})^{2}$
      & $-1.0$ \\

    \addlinespace
    Base‐height deviation
      & $r_{\mathrm{height}} = (z - z_{\mathrm{target}})^{2}$
      & $-10.0$ \\
    XY linear‐velocity tracking
      & $r_{\mathrm{trackLin}} = \exp\!\Bigl(-\|\dot P^b_{xy}-\mathrm{CMD}_{xy}\|^{2}/\sigma\Bigr)$
      & $1.0$ \\
    Yaw angular‐velocity tracking
      & $r_{\mathrm{trackAng}} = \exp\!\bigl(-(\omega^b_{z}-\mathrm{CMD}_{\mathrm{yaw}})^{2}/\sigma \bigr)$
      & $0.5$ \\

    \addlinespace
    Joint‐acceleration penalty
      & $r_{\mathrm{dofAcc}} = \sum_j\Bigl(\tfrac{\dot q_{t-1,j}-\dot q_{t,j}}{\Delta t}\Bigr)^{2}$
      & $-2.5\times10^{-7}$ \\
    Joint‐velocity penalty
      & $r_{\mathrm{dofVel}} = \sum_j(\dot q_{j})^{2}$
      & $-1\times10^{-3}$ \\
    Action‐rate penalty
      & $r_{\mathrm{actRate}} = \sum_j(a_{t-1,j}-a_{t,j})^{2}$
      & $-0.01$ \\

    \addlinespace
    Collision penalty
      & $r_{\mathrm{collision}} = \sum_i \mathbf{1}\bigl(\|\mathbf f_i\|>0.1\bigr)$
      & $0.0$ \\
    DOF position‐limit penalty
      & $r_{\mathrm{posLim}} = \sum_j \max\{\,0,\;\dots\}$
      & $-5.0$ \\
    No‐velocity contact penalty
      & $r_{\mathrm{contactNoVel}} = \sum_{i,j}(\text{contactVel}_{i,j})^{2}$
      & $-0.2$ \\

    \addlinespace
    Feet‐swing‐height penalty
      & $r_{\mathrm{feetSwing}} = \sum_j(\,h_{j}-0.08\,)^{2}\,\mathbf1(\text{in swing})$
      & $-20.0$ \\
    Alive bonus
      & $r_{\mathrm{alive}} = 1.0$
      & $0.15$ \\
    Hip‐position deviation
      & $r_{\mathrm{hipPos}} = \sum_{j\in\{1,2,7,8\}}q_{j}^{2}$
      & $-1.0$ \\

    \bottomrule
  \end{tabular}
\end{table}

\end{document}